\documentclass[conference]{IEEEtran}
\IEEEoverridecommandlockouts
\usepackage{cite}
\usepackage{amsmath,amssymb,amsfonts}
\usepackage{algorithmic}
\usepackage{graphicx}
\usepackage{textcomp}
\usepackage{xcolor}
\usepackage{float}
\usepackage{booktabs}
\usepackage{pgfplots}
\usepackage{caption}
\usepackage{stfloats}
\usepackage{listings}
\usepackage{multirow}
\usepackage{url}
\usepackage{longtable}

\def\BibTeX{{\rm B\kern-.05em{\sc i\kern-.025em b}\kern-.08em
    T\kern-.1667em\lower.7ex\hbox{E}\kern-.125emX}}
\begin{document} 

\title{Can LLM-assisted regularization increase forecast accuracy for migration flows in low data regimes?
}

\author{\IEEEauthorblockN{Nathaniel T. Hindman}
\IEEEauthorblockA{\textit{Artificial Intelligence} \\
\textit{Worcester Polytechnic Institute}\\
Worcester, Massachusetts \\
nthindman@wpi.edu}
\and
\IEEEauthorblockN{Fabricio Murai}
\IEEEauthorblockA{\textit{Data Science} \\
\textit{Worcester Polytechnic Institute}\\
Worcester, Massachusetts \\
fmurai@wpi.edu}
}

\maketitle

\begin{abstract}

Predicting migration flows remains a significant challenge for traditional gravity-based forecasting models, which primarily rely on structured socio-economic indicators such as economic disparity, political stability, and geographic distance. This work investigates whether Large Language Models (LLMs) can improve migration forecasting by extracting contextual migration-related signals from news articles and incorporating them into a weighted Lasso forecasting framework through feature-specific regularization penalties. The proposed framework uses hierarchical LLM inference pipelines to classify migration-related push--pull signals from news data and evaluates the resulting forecasting performance across multiple migration corridors between November 2021 and November 2022, including Mexico--United States, Ukraine--Poland, and Syria--Turkey.
Experimental results showed mixed performance across migration corridors and modeling strategies, and no single regularization approach consistently outperformed the others across all experiments. The best-performing Mexico configuration, which consisted of a gravity-based model augmented with the proposed push--pull ratios, achieved a Mean Absolute Percentage Error (MAPE) of 17.15\%, while the strongest Syria configuration achieved a MAPE of 29.29\% using Direct LLM-Lasso. For Ukraine, the best-performing configuration used LLM-Assisted Regularization (AR) and achieved a MAPE of 41.05\%. Overall, the results suggest that contextual article-derived features and LLM-guided regularization can improve migration forecasting under certain conditions, although migration corridor characteristics, article volume, and hyperparameter configuration strongly influenced performance.
\end{abstract}

\section{Introduction}

Human migration, particularly when conflict, economic instability, or environmental disruption drives it, presents a critical challenge for policymakers and humanitarian organizations. Accurate prediction of migration flows enables governments and aid organizations to allocate resources, plan infrastructure, and intervene more effectively in regions experiencing sudden population displacement.

Traditional migration forecasting approaches commonly rely on gravity-based modeling frameworks and structured socio-economic indicators such as economic disparity, geographic distance, political stability, demographic structure, and existing migrant networks. Prior research has shown that economic opportunity, inequality, diaspora effects, and socio-political instability strongly influence migration flows, while also highlighting the broader complexity and uncertainty associated with migration forecasting itself \cite{kluge2024migration}. Although gravity-based methods effectively model long-term migration relationships, they often fail to capture rapidly evolving real-world conditions reflected in unstructured information sources such as news media. Researchers have identified similar challenges in broader socio-economic forecasting and nowcasting literature, where many structured indicators are reported only monthly, quarterly, or annually and are frequently released with substantial publication delays, limiting their usefulness for real-time monitoring and forecasting applications \cite{barbaglia2023forecasting,hopp2022nowcasting}. These limitations have increased interest in high-frequency alternative data sources, including news media, social media, and other textual indicators, as supplementary contextual signals for forecasting and forced displacement modeling \cite{singh2019forcedmigration,iliopoulos2022pressmonitoring}.

Traditional migration forecasting also faces important data availability and temporal-resolution limitations. Many commonly used socio-economic indicators, including GDP per capita, Human Development Index (HDI), Global Peace Index (GPI), life expectancy, and education-related measures, are updated only annually or infrequently despite migration dynamics often evolving at monthly or even weekly timescales. In conflict-driven migration scenarios, instability, disrupted reporting infrastructure, and the difficulty of collecting reliable survey data in affected regions can delay, limit, or entirely prevent the availability of structured data. Prior work has noted that monitoring refugee displacement and return becomes particularly difficult in war-prone environments where systematic migration data remains limited or difficult to obtain, motivating the use of social media and textual data sources as supplementary contextual indicators for migration monitoring and forecasting \cite{WALK2023106268}.

Migration forecasting additionally suffers from limited availability of directly observed migration-flow data. Prior work has noted that researchers often derive migration estimates from census transition data collected over long temporal intervals, resulting in sparse temporal availability and substantial missing information during intercensal periods \cite{apariciocastro2024bayesian}. These limitations become particularly problematic in rapidly evolving migration scenarios in which socio-economic and geopolitical conditions may change substantially within short timeframes. As a result, supplementary high-frequency contextual information sources such as news media may provide useful signals for capturing migration-related dynamics that structured datasets do not immediately reflect. This study evaluates forecasting performance across a shared monthly window spanning November 2021 through November 2022 in order to maintain consistent temporal coverage across all evaluated migration corridors while capturing multiple migration conditions ranging from relatively stable migration behavior to conflict-driven volatility. The selected period additionally allowed evaluation during the emergence of the Ukraine conflict while preserving comparable forecasting conditions for the Mexico--United States and Syria--Turkey corridors.

Recent advances in Large Language Models (LLMs) and natural language processing systems have created new opportunities for extracting structured information from unstructured textual data. Prior studies have demonstrated that textual signals derived from news media, social media, and large-scale open-source datasets can function as early-warning indicators for migration forecasting and asylum prediction \cite{singh2019forcedmigration,iliopoulos2022pressmonitoring}. Prior work has additionally shown that LLMs can perform contextual text classification, sentiment analysis, and multi-label inference tasks across diverse textual domains while requiring substantially less manual feature engineering than traditional NLP pipelines \cite{wang2024adaptablereliabletextclassification}.

Rather than using LLMs as direct predictors, this work investigates whether contextual information extracted through LLM-assisted article classification can indirectly improve migration forecasting through feature weighting and regularization. Specifically, the proposed framework analyzes migration-related news articles to identify push--pull migration signals and incorporates semantic feature-category evidence computed from processed article text into a weighted Lasso framework.

The proposed framework combines structured gravity-model features with contextual information extracted from migration-related news articles in order to dynamically adjust feature importance based on evolving migration-related conditions. Rather than relying exclusively on static socio-economic indicators, the system incorporates contextual signals associated with political instability, economic disruption, conflict, and migration-related narratives extracted from news media.

Our main contributions are as follows:

\begin{itemize}
    \item We introduce an LLM-assisted regularization framework for migration forecasting that integrates structured gravity-model features with contextual information extracted from news articles.

    \item We evaluate multiple hierarchical LLM inference pipelines for extracting migration-related push--pull signals while balancing contextual understanding and computational cost.

    \item We compare direct LLM-generated feature penalties against a more stable count-based weighting approach derived from aggregated semantic feature-category evidence extracted through prompt-guided LLM interpretation of processed article corpora.

    \item We evaluate the proposed framework across multiple migration corridors, including Mexico--United States, Ukraine--Poland, and Syria--Turkey, using sliding  monthly forecasting experiments and MAPE evaluation.
\end{itemize}

Our experimental results show that incorporating news-derived contextual signals can improve predictive performance in certain migration corridors, particularly when sufficient article coverage exists. Across several experiments, gravity models augmented with push--pull ratio features consistently outperformed direct LLM-generated penalties, while count-based LLM-assisted regularization produced more stable behavior than direct LLM-Lasso approaches. Results further suggest that the usefulness of article-derived weighting depends strongly on both article quantity and the quality of extracted signals.

\section{Background and Related Work}

We review the three most closely related body of works, namely: traditional gravity-based models for migration flows, the use of LLMs for L1-regularization, and cross-modality models that make use of news text for predicting migration flows. 

\subsection{Gravity Models of Migration}
Migration forecasting has traditionally relied on gravity-based modeling frameworks that estimate population movement between origin and destination regions using structured socio-economic and geographic variables \cite{caballero2024gravity}. Inspired by Newtonian gravity models, migration flow is generally assumed to increase with the ``mass'' of origin and destination populations while decreasing as geographic, political, or economic barriers increase. Due to their interpretability, theoretical grounding, and flexibility in incorporating diverse structural indicators, gravity-based models have been widely used in international migration forecasting \cite{caballero2024gravity}.

Researchers commonly model migration flows using gravity formulations \cite{cutuli2024bayesianhierarchical}. In the following formulation, \(M_{i,j,t}\) represents migration flow from origin \(i\) to destination \(j\) at time step \(t\), while \(P_{i,t}\) and \(P_{j,t}\) represent the population sizes of the origin and destination regions, respectively. The variables \(X_{k,t}\) denote additional explanatory features such as economic indicators, political stability measures, development metrics, or geographic factors. The coefficients \(\beta_k\) capture the contribution of each feature to migration flow, and \(\epsilon\) represents residual error. This work uses the following generalized gravity-based migration formulation
\begin{equation*}\label{eq:gravity_base}
\log M_{i,j,t} = \beta_0 
+ \beta_1 \log P_{i,t} 
+ \beta_2 \log P_{j,t} 
+ \sum_{k=3}^m\beta_kX_{k,t}
+ \epsilon ,
\end{equation*}
where migration flow is modeled as a function of both population-based gravity effects and additional structured explanatory variables. This generalized formulation allows gravity-based models to flexibly incorporate a wide range of socio-economic, political, demographic, and geographic indicators depending on the migration corridor and forecasting objective being studied.
 
Gravitation models vary by application and there is no universally applicable gravitation formula\cite{poot2016gravity}.  Each different migration model is built to cater to specific features and circumstances to better encapsulate the behavior of specific country pairs of interest\cite{beyer2022gravity}.  

\subsection{L1 Regularization, LLM-Lasso and Feature Weighting}
In machine learning, L1 regularization is commonly used to control model complexity by penalizing the absolute values of model coefficients, often driving less important features toward zero. The standard L1-regularized objective function can be expressed as
\begin{equation} \label{eq:lasso}
\min_{\beta} \;\; \|y - X\beta\|^2 + \lambda \sum_{j=1}^{p} |\beta_j|,
\end{equation}
where \(\lambda\) controls the strength of regularization. This regularization approach encourages sparse model representations by reducing the influence of less informative features while retaining the most predictive variables. As the regularization strength \(\lambda\) increases, a larger number of feature coefficients are driven toward zero, resulting in simpler and more constrained models.

LLM-Lasso extends standard L1 regularization by incorporating feature-specific penalty weights derived from language model outputs \cite{zhang2025llmlasso}. In this formulation, each coefficient \(\beta_j\) is penalized according to a feature-specific weight \(w_j\), allowing more important features to receive lower penalties while less relevant features are more strongly regularized. The resulting objective function becomes
\begin{equation} \label{eq:lasso_with_wj}
\min_{\beta} \;\; \|y - X\beta\|^2 + \lambda \sum_{j=1}^{p} w_j |\beta_j|,
\end{equation}
where \(w_j\) is the penalty weight associated with feature \(j\).

While prior work relies on directly generated LLM weights, these approaches can be unstable. This motivates alternative approaches for estimating penalty weights.

 \subsection{Incorporating News Data into Migration Forecasting}

Recent advances in Large Language Models (LLMs) and natural language processing systems have significantly expanded the capabilities of automated text classification and contextual reasoning systems. Prior work has demonstrated that LLMs can effectively perform sentiment analysis, multi-label classification, contextual categorization, and reasoning tasks across diverse textual domains while requiring substantially less manual preprocessing and feature engineering than traditional NLP pipelines. Although the proposed framework shares similarities with retrieval-augmented gen- eration (RAG) systems, it does not implement a traditional embedding-based vector retrieval architecture.\cite{wang2024adaptablereliabletextclassification}.

Recent studies have additionally explored the use of unstructured textual data and open-source information sources for migration forecasting and displacement modeling. Singh et al. \cite{singh2019forcedmigration} combined traditional migration indicators with signals derived from social media and newspaper data to model forced migration dynamics, demonstrating that open-source textual signals can improve predictive performance relative to traditional variables alone. Similarly, Iliopoulos et al. \cite{iliopoulos2022pressmonitoring} proposed a migration forecasting framework based on national press monitoring, topic modeling, and neural network architectures using GDELT and EUROSTAT data as early-warning indicators for asylum application forecasting in Europe. Additional migration forecasting work has further explored the integration of textual signals and event-driven information into predictive migration frameworks \cite{carammia_forecasting_2022}.  The usefulness of textual and multimodal contextual information has also been demonstrated in other forecasting domains. In transportation forecasting, prior studies have shown that integrating social media signals and agency-generated textual feeds with historical traffic measurements can improve predictive performance  \cite{liu2024lingkai,zhang2024chattraffic}.

These developments suggest that LLMs may be well-suited for identifying migration-related contextual signals within news articles. Migration events are frequently associated with political instability, economic hardship, conflict, humanitarian crises, labor opportunities, and climate-related disruption, many of which are discussed qualitatively within news media before becoming fully reflected in structured numerical datasets \cite{qi2023forcedmigration,singh2019forcedmigration}. Because LLMs are capable of contextual text classification, reasoning, and sentiment-oriented interpretation, they provide a potential mechanism for identifying whether migration-related narratives within articles correspond more closely to push-oriented, pull-oriented, mixed, or neutral migration pressures.

Motivated by these observations, the proposed framework utilizes LLMs to analyze migration-related news articles and extract migration-related push--pull signals from unstructured textual data. Rather than using LLMs as direct predictors of migration flow, the system employs them primarily as contextual inference mechanisms for article classification and signal extraction. In addition to push--pull classification, the framework computes aggregated semantic feature-category evidence using predefined feature descriptions and representative migration-related indicators embedded within the LLM prompt. The framework subsequently incorporates these extracted semantic signals into gravity-based forecasting models through push–pull ratio features and feature-specific regularization techniques.

\section{Push-Pull Features for Gravitation Models and LLM-Assisted Regularization}

This work proposes a hybrid framework that integrates structured gravity-based migration forecasting with contextual signals extracted from unstructured news data using LLMs (LLMs). The proposed system consists of two primary components: a hierarchical inference framework for extracting migration-related push--pull signals from news articles, and a weighted Lasso forecasting model in which feature penalties are informed by aggregated semantic feature-category evidence derived from the processed article corpus.

Rather than relying on LLMs to directly predict migration flows or assign migration coefficients, the framework uses LLM-assisted article classification together with aggregated semantic feature-category evidence to guide feature regularization. The key idea is that news articles capture contextual information about migration-related conditions occurring within a given time period. Prior work has shown that combining traditional migration indicators with open-source textual and media-derived signals can improve forced migration forecasting performance and provide useful early-warning indicators for displacement modeling \cite{singh2019forcedmigration}.

By transforming aggregated semantic feature-category evidence into feature-specific penalty weights, the model can dynamically adjust the importance of predictors based on contextual signals extracted from migration-related news data. This formulation allows the framework to retain the interpretability of traditional gravity models while incorporating time-sensitive information derived from unstructured textual sources.

\subsection{Proposed features: push-pull count ratios}

The proposed model incorporates a set of socio-economic, demographic, and news-derived features to capture the primary drivers of migration flows. The model represents economic conditions using gross domestic product per capita ($\text{GDP}$) and gross national income per capita ($\text{GNI}$), capturing relative wealth and economic opportunity. The model captures development and quality of life using the Human Development Index ($\text{HDI}$), along with supporting indicators including life expectancy at birth and expected years of schooling. The model captures political and conflict-related conditions through the Global Peace Index ($\text{GPI}$), which reflects levels of stability and security within a country. The framework incorporates these features for both origin ($i$) and destination ($j$) countries, allowing the model to capture asymmetries in migration incentives, and are complemented by geographic factors such as distance and corridor-specific effects.

In addition to structured country-level indicators, the model incorporates migration-related signals derived from news media. The framework classifies each article into one of four categories: push, pull, mixed, or unclear. These classifications are aggregated at the monthly level to construct normalized ratio-based features. For a given month $t$, the total number of articles is denoted by $N_t$, and the push, pull, mixed, and unclear ratios are defined as the proportion of articles assigned to each category. For example, the push ratio is computed as $\textsc{PushRatio}_t = \textsc{PushCount}_t/{N_t}$, with analogous definitions for $\textsc{PullRatio}_t$, $\textsc{MixedRatio}_t$, and $\textsc{UnclearRatio}_t$.

In practice, the mixed and unclear categories constitute only a small fraction of the total article volume in most months, with the majority of articles being classified as either push or pull. As a result, the corresponding ratio features typically take on values close to zero and serve primarily as minor adjustment terms rather than dominant drivers. Overall, these ratio-based features provide a normalized representation of the distribution of migration-related signals in the news, allowing the model to capture the relative prevalence of push- and pull-driven narratives while remaining robust to variation in total article volume across time. One limitation of the proposed framework is that both the push--pull ratio features and the feature-specific regularization penalties are derived from the same underlying article corpus. Although these components serve different purposes within the forecasting framework---with push--pull ratios representing overall migration-related directional pressure and the regularization process influencing feature-specific weighting during optimization---some degree of shared contextual information may still exist between the two representations.
\subsection{The feature-augmented gravitational model}
While gravitational methods are commonly used to model migration, the proposed framework extends traditional gravity-based formulations by incorporating both structured socio-economic indicators and contextual features extracted from migration-related news media. Using the previously defined features, the proposed migration model can be expressed as
\begin{equation*}
\begin{aligned}
\log\!\left(1+M_{i,j,t}\right) =\;&
\beta_0 
+ \beta_1 \log\!\left(1+\text{GDP}_{i,t}\right) \\
&+ \beta_2 \log\!\left(1+\text{GDP}_{j,t}\right)
+ \beta_3 \log\!\left(1+\text{GNI}_{i,t}\right) \\
&+ \beta_4 \log\!\left(1+\text{GNI}_{j,t}\right)
+ \beta_5 \log\!\left(1+D_{i,j}\right) \\
&+ \beta_6 \,\text{HDI}_{i,t}
+ \beta_7 \,\text{HDI}_{j,t} \\
&+ \beta_8 \,\text{GPI}_{i,t}
+ \beta_9 \,\text{GPI}_{j,t} \\
&+ \beta_{10} \,\text{LifeExp}_{i,t}
+ \beta_{11} \,\text{LifeExp}_{j,t} \\
&+ \beta_{12} \,\text{ExpSchool}_{i,t}
+ \beta_{13} \,\text{ExpSchool}_{j,t} \\
&+ \beta_{14} \,\text{PushRatio}_t
+ \beta_{15} \,\text{PullRatio}_t \\
&+ \beta_{16} \,\text{MixedRatio}_t
+ \beta_{17} \,\text{UnclearRatio}_t \\
&+ \alpha_{i,j}
+ \epsilon_{i,j,t}.
\end{aligned}
\end{equation*}
where $M_{i,j,t}$ denotes migration flow from origin country $i$ to destination country $j$ at time step $t$, $\alpha_{i,j}$ represents corridor-specific effects, and $\epsilon_{i,j,t}$ denotes residual error.

\subsection{Proposed method: LLM-Assisted Regularization}
In traditional L1-regularized models, all features are penalized uniformly, regardless of their contextual relevance (eq.~\ref{eq:lasso}). However, in migration modeling, the importance of features such as economic conditions, conflict, or political stability can vary significantly depending on current events. However, due to limitations on historical data on migration flows, it is extremely difficult to learn their relative importance directly from data via hyperparameter tuning. To address this limitation, we propose a method for determining feature-specific penalty weights based on their relevance within a corpus of migration-related articles (eq.~\ref{eq:lasso_with_wj}).

Rather than relying on LLMs to directly assign numerical importance scores---which can be unstable or inconsistent---we instead derive penalty weights from aggregated semantic feature-category evidence computed over the processed article corpus. We associate each predefined feature category with representative migration-related indicators and contextual descriptions embedded within the LLM prompt. For each feature category \(j\), the framework computes a semantic evidence count through prompt-guided LLM interpretation using predefined feature descriptions and representative migration-related indicators associated with that category (Appendix D). The framework subsequently transforms these counts into feature-specific smooth penalty weights through an inverse relationship given by
\[
w_j = \frac{1}{1 + \mathrm{count}_j},
\]
where \(\mathrm{count}_j\) denotes the aggregated amount of semantic evidence identified for feature category \(j\) through prompt-guided contextual interpretation across the processed article corpus. We selected this inverse weighting formulation primarily because it provides a simple and interpretable mechanism for reducing regularization pressure on frequently referenced migration-related features while avoiding excessively large weight magnitudes. The formulation additionally limits the role of the LLM to contextual evidence identification rather than requiring the model to directly generate numerical feature coefficients or importance scores directly. The prompt used for colecting feature counts is listed in (Appendix~\ref{appendix:Feature-Category Counting Prompt for LLM-Assisted Regularization}).

\subsection{Extracting Push-Pull Factors from News Media with LLMs}

A central component of the proposed framework is the extraction of migration-relevant signals from unstructured news data. To achieve this, we employ a hierarchical LLM-based pipeline designed to classify articles while balancing computational efficiency and inference quality.

Preliminary analysis indicated that larger language models generally produce more reliable push--pull classifications than smaller models, but at a significantly higher computational cost. Recent work on language model cascades and hierarchical inference strategies has explored frameworks in which lightweight models process simpler inputs while more computationally expensive models are selectively invoked only for more difficult or ambiguous cases \cite{gupta2024cascades,varshney2022model}. These approaches aim to balance inference quality and computational efficiency through adaptive multi-stage routing. Motivated by similar trade-offs, the proposed framework employs a staged inference pipeline in which lightweight models are applied first, while larger models or full-article analysis are used only when additional contextual reasoning is required.

In the proposed pipeline, each article is initially analyzed at the URL level (typically the title), where a lightweight model attempts to classify it into one of four categories: push, pull, mixed, or neutral (Appendix~\ref{appendix:URL-Level Classification Prompt}).  If the classification is uncertain, the pipeline escalates either to a higher-capacity model or to full article text in order to incorporate additional context (Appendix~\ref{appendix:LLM Prompt for Article-Based Push--Pull Classification}). We consider multiple hierarchical configurations of this process, which differ in how model capacity and textual context are combined.

The naming convention used throughout this work reflects both the model scale and the level of textual context being analyzed. The term ``Su'' (Small--URL) refers to a lightweight language model processing URL-level text, typically consisting of the article headline extracted from the URL. ``Lu'' (Large--URL) refers to a larger-capacity language model processing the same URL-level information in cases where the smaller model is unable to confidently classify the article. Finally, ``Sa'' (Small--Article) refers to a lightweight model processing the full article text to incorporate broader contextual information when URL-level inference alone is insufficient.

We consider three possible pipelines:
\begin{itemize}
    \item SuSa (Small--URL, Small--Article): applies Qwen 2.5 7B to the URL text and escalates to full article processing using the same model if the classification remains uncertain. This configuration evaluates whether additional contextual information alone is sufficient to improve classification quality without increasing model size.
    \item SuLu (Small--URL, Large--URL): instead escalates from Qwen 2.5 7B to Qwen 2.5 72B while continuing to rely exclusively on URL-level text. This approach evaluates whether increased model capacity alone can resolve ambiguity without requiring access to the full article.
    \item SuLuSa (Small--URL, Large--URL, Small--Article): combines both escalation strategies within a multi-stage inference pipeline. Qwen 2.5 7B first analyzes the URL text to produce an initial push--pull classification. If the result remains uncertain, Qwen 2.5 72B is subsequently applied to the same URL-level information. If ambiguity persists after this second stage, the pipeline escalates to full article processing using Qwen 2.5 7B. This design combines lightweight inference, larger-model reasoning, and contextual expansion in a hierarchical manner intended to balance inference cost with classification reliability.
\end{itemize}

\subsection{Forecasting Configurations}

To evaluate the effectiveness of article-derived contextual signals within migration forecasting, we compare four primary forecasting configurations:

\begin{itemize}

\item \textbf{Gravity with Proposed Push/Pull Ratios (Gravity-PP):}
A gravity-based forecasting model augmented with the proposed normalized push, pull, mixed, and unclear article-ratio features derived from migration-related news articles. These ratio features provide a structured representation of migration-related contextual signals extracted from the article corpus.

\item \textbf{Proposed LLM-Assisted Regularization (LLM-AR):}
The proposed count-based weighted regularization framework in which feature-specific penalties are derived indirectly from aggregated semantic feature-category evidence computed from the processed article corpus. The regularization process therefore assigns smaller penalties to frequently referenced migration-related features during regularization.

\item \textbf{Gravity Model:}
A baseline gravity-based migration forecasting model using structured socio-economic and geographic indicators together with standard L1-regularization. This configuration serves as the primary control model throughout the study.

\item \textbf{Direct LLM-Lasso:} In this formulation, the LLM analyzed migration-related news articles associated with each forecasting period and directly assigned feature-specific regularization weights \(w_j\) based on the inferred importance of socio-economic, political, and migration-related variables. This approach closely follows the general methodology proposed in the original LLM-Lasso framework, where LLMs are used to generate domain-informed feature penalties for weighted L1-regularization \cite{zhang2025llmlasso}. We include Direct LLM-Lasso as a comparative baseline. The Direct LLM-Lasso framework uses a structured prompt to assign feature-specific monthly regularization penalties directly from article evidence (Appendix~\ref{appendix:Direct LLM-Lasso Prompt}).

\end{itemize}

The following experiments compare these forecasting configurations across multiple migration corridors, article volumes, and regularization settings using sliding  monthly forecasting evaluation and MAPE.

\subsection{Migration Flow Datasets}

We evaluate the models using monthly migration-flow observations obtained from the Meta International Migration Flows dataset. Specifically, we obtain data for the Mexico--United States, Ukraine--Poland, and Syria--Turkey migration corridors corresponding to the period from November 2021 to November 2022. Some basic statistics about each corridor are shown in Table~\ref{tab:flows_stats}.

\subsection{News Datasets}
The Global Database of Events, Language, and Tone (GDELT) project is a large-scale open-access platform that monitors worldwide news coverage and extracts structured event, theme, and location information from global media sources. Prior migration forecasting research has utilized GDELT together with NLP pipelines and topic modeling for early-warning prediction of asylum and migration flows \cite{iliopoulos2022pressmonitoring}.

During early-stage pipeline development and evaluation, this study utilized GDELT as a source of migration-related and conflict-related news data while evaluating hierarchical inference configurations such as SuSa, SuLu, and SuLuSa. However, inconsistent coverage and limited article availability across several migration corridors limited its usefulness for large-scale forecasting experiments. To obtain larger and more consistent article datasets, the study later transitioned toward collecting full-text articles directly from mainstream news sources including PBS, NPR, NBC, CBS, and BBC.

The final dataset consists of migration-related articles associated with three migration corridors: Mexico--United States, Ukraine--Poland, and Syria--Turkey. We collected and grouped articles at the monthly level in order to align with the sliding migration forecasting framework used throughout the study.

We collected articles for monthly forecasting windows spanning November 2021 through November 2022 using query-based web scraping and article retrieval pipelines. For each migration corridor, search queries combined region-specific terms and migration-related event keywords with source-specific domain filters in order to retrieve relevant news coverage from mainstream media outlets. The scraping pipeline retrieved candidate articles using DuckDuckGo Search (DDGS), after which the pipeline extracted full-text article content using Newspaper3k with a BeautifulSoup-based fallback extraction pipeline when necessary. The pipeline de-duplicated retrieved articles and grouped them by month and migration corridor, and balanced across sources and months prior to downstream LLM-based processing and forecasting experiments. Additional scraping queries and retrieval templates are provided in Appendix~\ref{appendix:article_queries}.

\subsection{Model Training}

The forecasting framework uses a sliding nine-month training window in which the model is trained on the previous nine months of migration-flow data and evaluated on the following holdout month. We selected a nine-month window in order to balance historical context with adaptability to rapidly evolving migration conditions and conflict-driven migration spikes. Shorter windows risked insufficient historical information, while substantially longer windows could reduce the model's ability to adapt to temporal shifts in migration behavior. By continuously retraining on more recent observations while discarding older data, the framework can better adapt to evolving migration patterns that may not generalize well across longer fixed training periods~\cite{forecastingeval2022}. Prior to the availability of article data, the framework operated as a standard gravity-based Lasso model using uniform regularization across structured migration and country-level features. Once article data became available for November 2021, article-derived features and feature-specific $w_j$ regularization weights were incorporated into the training process through the proposed LLM-assisted regularization framework.

\begin{table}[t]
\centering
\caption{Summary statistics for migration-flow datasets across evaluated migration corridors.}
\label{tab:flows_stats}
\setlength{\tabcolsep}{5pt}

\begin{tabular}{lrrrr}
\toprule
\textbf{Corridor} & \textbf{Min} & \textbf{Max} & \textbf{Mean} & \textbf{Std. Dev.} \\
\midrule
Mexico--US & 8,718 & 102,939 & 46,194 & 23,840 \\
Ukraine--Poland & 998 & 163,937 & 18,628 & 30,839 \\
Syria--Turkey & 1,959 & 17,395 & 5,821 & 3,325 \\
\bottomrule
\end{tabular}

\end{table}


\section{Experiments}
This section describes two main sets of experiments used for evaluating the contributions of the paper.

\subsection{Part I: Extracting Push--Pull Factors}

This set of experiments evaluates hierarchical LLM inference pipelines designed to classify migration-related news articles into push--pull migration categories. The experiments focus on comparing different hierarchical inference strategies, model architectures, and temperature settings while using a shared migration-related article corpus.

\subsubsection*{Evaluation Setup}

We evaluated hierarchical inference pipelines using a dataset of 381 migration-related articles sourced from GDELT. All evaluated configurations processed the same article corpus in order to allow controlled comparison across hierarchical inference strategies and model architectures.

Push--pull classification behavior was evaluated using the number of articles classified into the categories ``push,'' ``pull,'' ``mixed,'' ``neutral,'' and ``unclear.'' Lower proportions of ``unclear'' classifications were treated as indicators of stronger contextual classification confidence. Because comprehensive manual validation across the full dataset was outside the scope of this study, we additionally manually reviewed approximately 20--40 article classifications across multiple configurations as a limited qualitative evaluation of classification behavior.

\subsubsection*{Pipeline Configurations}

We evaluated four hierarchical inference configurations: Qwen SuSa, Qwen SuLu, Qwen SuLuSa, and a Mistral-based SuSa configuration. Each pipeline used different combinations of lightweight URL-level inference, larger-model escalation, and full-article contextual analysis in order to balance computational efficiency and contextual reasoning capability. 

\subsubsection*{Temperature Evaluation}

We evaluated all inference pipelines across temperature settings $\tau \in \{0.1, 0.3, 0.5, 0.8, 0.9, 1.5\}$. Each pipeline was executed on the same article corpus at every temperature setting in order to evaluate classification consistency under varying levels of generation randomness. Within Ollama, temperature values range from 0 to 2, where lower values produce more deterministic outputs and higher values increase response variability and randomness. Because a temperature of 0.8 is the default setting in Ollama, it was treated as the control configuration for comparison in order to evaluate the effects of both lower-temperature deterministic inference and higher-temperature randomized generation behavior across the evaluated pipelines.


\subsection{Part II: LLM-based Regularization Approaches}

This set of experiments evaluates the effectiveness of incorporating article-derived contextual information into gravity-based migration forecasting models through multiple feature-weighting and regularization strategies. The experiments compare traditional gravity-based forecasting against forecasting approaches augmented with migration-related signals extracted from news media using the proposed hierarchical inference framework.

This set of experiments evaluates the effectiveness of incorporating article-derived contextual information into gravity-based migration forecasting models through multiple feature-weighting and regularization strategies. The experiments compare traditional gravity-based forecasting against forecasting approaches augmented with migration-related signals extracted from news media using the proposed hierarchical inference framework.

\subsubsection*{Forecasting Experiments}

We conducted a series of forecasting experiments to evaluate both LLM-informed feature regularization and the incorporation of normalized push--pull article ratios as contextual forecasting features within gravity-based migration models. The evaluated forecasting configurations are described in Appendix~\ref{appendix:Best_Model_for_each_country_Pair}.

We conducted forecasting experiments across the Mexico--United States, Syria--Turkey, and Ukraine--Poland migration corridors using sliding monthly forecasting windows spanning November 2021 through November 2022. For each prediction month, the training window shifted forward by one month in order to continuously retrain on recent observations while evaluating predictive performance on future holdout periods. This rolling-window strategy allowed the forecasting models to adapt to evolving migration conditions while preserving temporal separation between training and evaluation periods in order to reduce temporal leakage \cite{forecastingeval2022}.

We evaluated multiple article-corpus sizes in order to examine whether increasing the quantity of migration-related news articles improved downstream forecasting performance. Initial experiments utilized a fixed corpus size of $N=100$ articles per migration corridor distributed across the 12-month forecasting period, corresponding to approximately 8--9 articles per month. We then conducted additional experiments using the maximum number of articles that could be collected for each corridor within the selected time period, resulting in approximately $N=1245$ articles for Mexico--United States, $N=1526$ articles for Ukraine--Poland, and $N=882$ articles for Syria--Turkey. These experiments were designed to compare relatively small, controlled article corpora against substantially larger article collections in order to evaluate whether increased news coverage produced measurable forecasting improvements. Intermediate corpus sizes were not extensively evaluated due to computational and time constraints.  Most collected news articles contained approximately 4{,}000 characters of text on average. For computational consistency during local inference, the first 3{,}000 characters of each article were supplied to the LLM during classification and feature-extraction stages.

We evaluated Lasso regularization strengths across sliding forecasting windows using $\lambda \in \{7\times10^{-4}, 9\times10^{-4}, 1\times10^{-3}, 3\times10^{-3}, 5\times10^{-3}, 6\times10^{-3}, 7\times10^{-3}, 8\times10^{-3}\}$. Final experiments primarily utilized $\lambda = \{1,3,5,8\}\times10^{-3}$ together with the SuLuSa hierarchical inference pipeline using Qwen 2.5 7B and Qwen 2.5 72B models. Article classification used temperature $0.2$, while feature-count and penalty-generation prompts used temperature $0.0$ in order to encourage deterministic outputs.

We evaluated forecasting performance primarily using Mean Absolute Percentage Error (MAPE) across held-out monthly migration predictions.

\section{Results}


\subsection{Part I results}

\subsubsection*{Results} Results shown in Table~\ref{tab:temp08_results} indicate that the SuLuSa configuration achieved the strongest overall performance by producing the lowest proportion of ``unclear'' classifications across most evaluated temperature settings. We attribute SuLuSa's advantage over the other pipelines to its multi-stage escalation strategy selectively ability to incorporate both increased model capacity and deeper contextual information when necessary. In contrast, the SuLu configuration generally performed worse because larger models operating solely on URL-level information lacked broader article context. These findings highlight the importance of contextual expansion in migration-related push--pull classification tasks and motivated our selection of SuLuSa as the primary pipeline for subsequent migration forecasting experiments.

\begin{figure}[tbp]
    \centering
    \includegraphics[width=\linewidth]{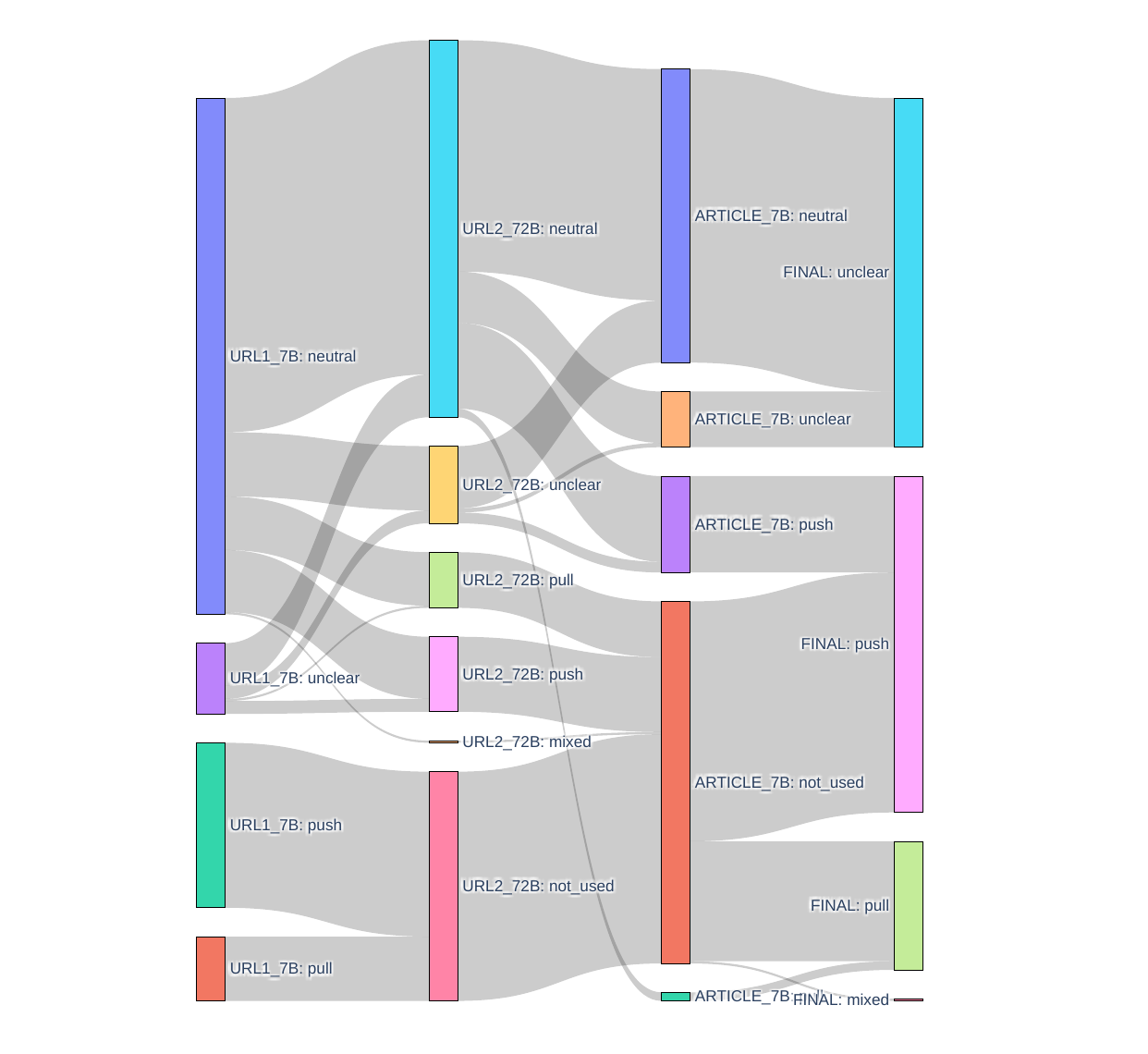}
    \caption{SuLuSa Classification across steps}
    \label{fig:mode3_sankey}
\end{figure}

\begin{figure}[tbp]
    \centering
    \includegraphics[width=0.7\linewidth]{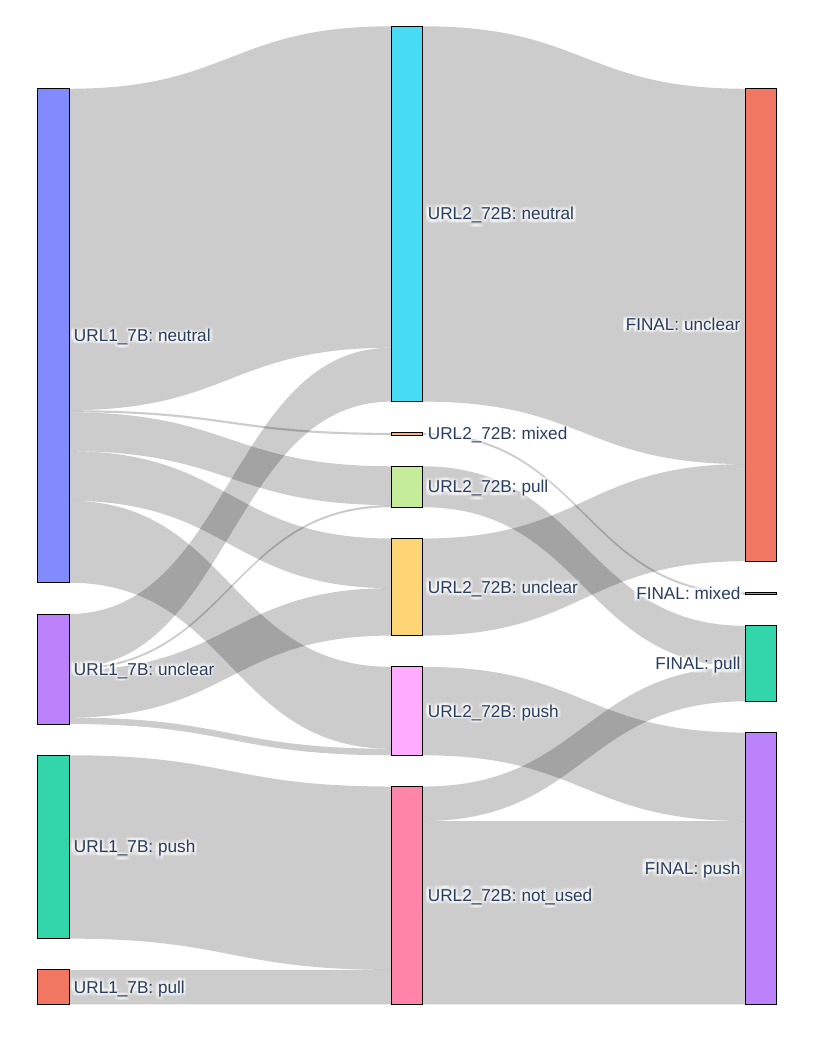}
    \caption{SuLu Classification across steps }
    \label{fig:mode2_sankey}
\end{figure}

Figures~\ref{fig:mode3_sankey} and~\ref{fig:mode2_sankey} illustrate the flow of article classifications across the hierarchical inference stages for the SuLuSa and SuLu pipelines, respectively. For SuLuSa, the addition of the final Sa (Small–Article) stage allows uncertain URL-level classifications to be resolved using full-article contextual information, reducing the number of samples that remain classified as ``unclear'' in the final output. In contrast, SuLu, which relies exclusively on URL-level reasoning even after escalation to a larger model, retains a larger proportion of unresolved classifications. These results suggest that incorporating additional contextual depth through full-article analysis contributes more substantially to classification quality than increasing model size alone.

These findings highlight the importance of incorporating full-text contextual understanding in migration-related push--pull classification tasks, as well as the effectiveness of hierarchical inference strategies that balance computational cost with contextual depth. Based on these results, SuLuSa was selected as the primary configuration for subsequent experiments involving downstream feature weighting and migration modeling. While SuLuSa consistently produced the strongest overall performance among the evaluated configurations, it also demonstrated relatively stable classification behavior across repeated evaluation runs on the same article corpus. 

\subsubsection*{Impact of Temperature and LLM Choice}

We evaluated inference pipelines across multiple temperature settings $\tau \in \{0.1, 0.3, 0.5, 0.8, 0.9, 1.5\}$. For each temperature value, we executed all four pipelines on the same dataset, allowing for a controlled comparison of both hierarchical structure and generation variability. We observe that temperature had minimal impact on overall classification behavior. For example, the Qwen SuLuSa inference pipeline classified 163 articles as ``unclear'' at a temperature of $\tau = 0.3$, compared to 159 articles at $\tau = 1.5$, demonstrating only minor variation across temperature settings (see Tables~\ref{tab:temperature} and \ref{tab:temp_sensitivity} in Appendix~\ref{app:temperature}). Although higher temperature settings introduced slightly greater output variability, the constrained prompt structure and fixed classification schema still restricted the range of allowable responses, which likely contributed to the relatively consistent behavior observed across temperature values.

\subsection{ Part II Results}

Table~\ref{tab:exp4_results} shows that forecasting performance varied across migration corridors and regularization strategies. Different configurations produced the strongest performance under different migration settings, with no single regularization approach consistently outperforming competing methods across all evaluated corridors. 
Direct LLM-Lasso produced competitive forecasting performance in several experiments, while the proposed LLM-Assisted Regularization (AR) framework achieved the strongest performance in others. These results indicate that the effectiveness of article-derived regularization strategies depends strongly on migration-corridor characteristics and forecasting conditions.

\begin{table}[tbp]
\centering
\caption{Inference Pipeline Performance at Temperature $\tau = 0.8$}
\label{tab:temp08_results}

\setlength{\tabcolsep}{4pt}

\begin{tabular}{lccccc}
\toprule
\textbf{Mode} & \textbf{Push} & \textbf{Pull} & \textbf{Mixed} & \textbf{Neutral} & \textbf{Unclear} \\
\midrule
Qwen SuSa     & 162 & 24 & 0  & 0 & 195 \\
Qwen SuLu     & 126 & 35 & 1  & 0 & 219 \\
Qwen SuLuSa   & 157 & 60 & 1  & 0 & \textbf{163} \\
Mistral SuSa  & 150 & 40 & 15 & 0 & 172 \\
\bottomrule
\end{tabular}

\end{table}

\begin{table}[t]
\centering
\caption{Direct LLM-Lasso performance compared to control and best gravity-based forecasting configurations (MAPE \%)}
\label{tab:exp4_results}
\begin{tabular}{l c c c c}
\toprule
 & \multirow{2}{*}{\textbf{Gravity}} & \textbf{Best} & \textbf{Direct} & \multirow{2}{*}{\textbf{LLM-AR}} \\
 &  & \textbf{Gravity-PP} & \textbf{LLM-Lasso} &  \\
\midrule
Mexico & 17.44 & \textbf{17.15} & 23.48 & 23.92 \\
Ukraine & 61.66 & 42.10 & 43.61 & \textbf{41.05} \\
Syria & 31.44 & 32.74 & \textbf{29.29} & 32.26  \\
\bottomrule
\end{tabular}
\end{table}

\subsubsection*{Impact of number of articles}

\begin{figure}[tbp]
    \centering
    \includegraphics[width=\linewidth]{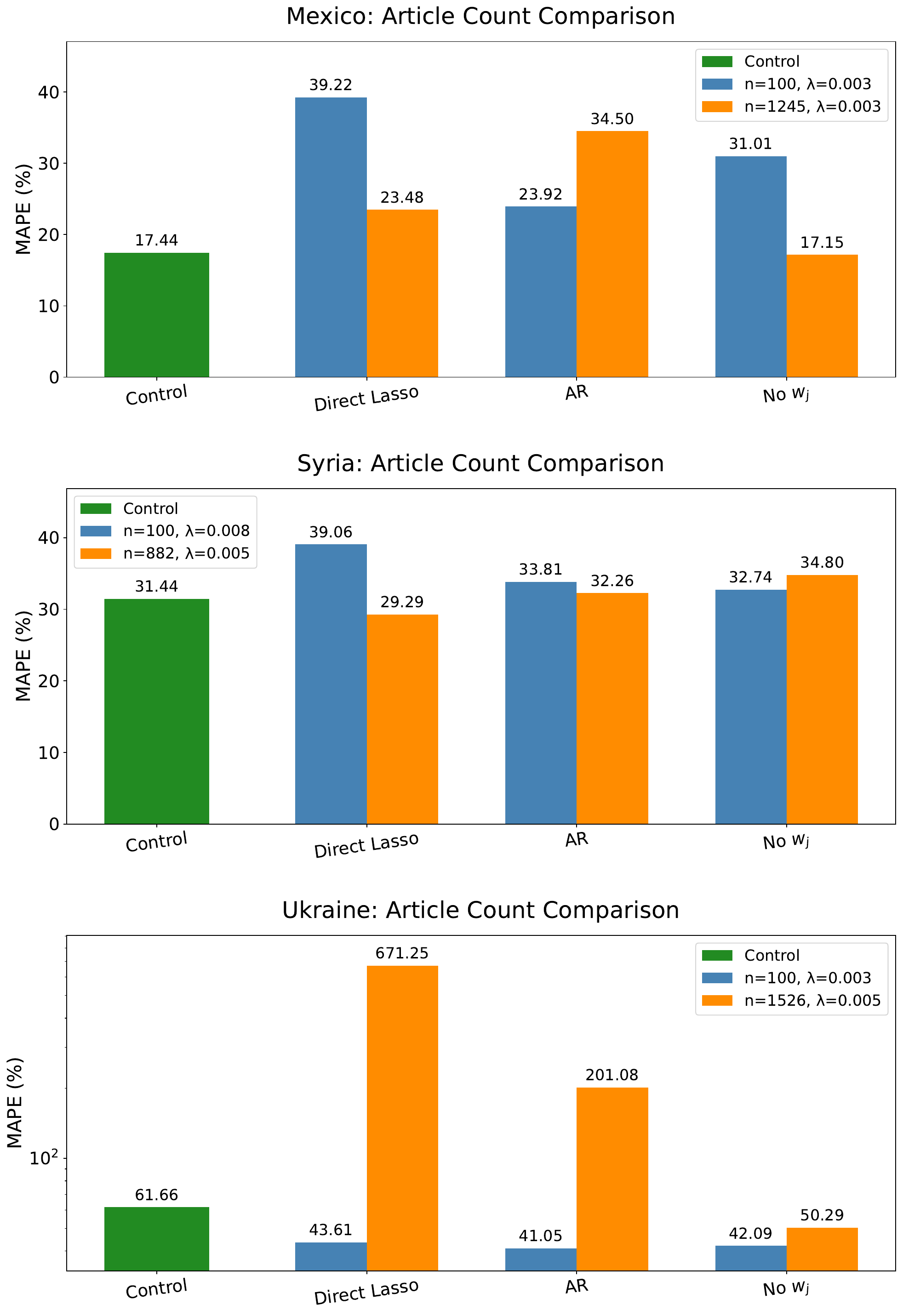}
    \caption{N-comparison across Models}
    \label{fig:N_comparison}
\end{figure}

Figure~\ref{fig:N_comparison} indicates that increasing the number of articles did not consistently improve predictive performance for either Direct LLM-Lasso or LLM-Assisted Regularization (AR). Across the evaluated migration corridors, the impact of larger article corpora varied substantially between modeling approaches and country pairs. Direct LLM-Lasso demonstrated improved performance for the Mexico--United States and Syria--Turkey corridors, but this trend did not generalize to the Ukraine corridor, where predictive performance remained inconsistent. 

Similarly, LLM-Assisted Regularization (AR) was frequently negatively affected by larger article volumes across several evaluated configurations. The standard Lasso formulation without feature-specific \(w_j\) regularization also produced mixed behavior, improving performance for the Mexico--United States corridor while producing weaker results for the Syria and Ukraine migration corridors. These findings suggest that increasing article volume alone does not guarantee improved forecasting accuracy and may introduce additional variability depending on the migration context and regularization strategy employed. 

Table~\ref{tab:exp5_results} summarizes forecasting performance across the evaluated migration corridors. Forecasting performance varied across forecasting configurations, with different regularization approaches producing the lowest forecasting error for different migration settings.

For the Mexico corridor, the strongest overall performance was achieved by the gravity model augmented with push--pull ratio features without feature-specific \(w_j\) regularization, which achieved a MAPE of \(17.15\%\), slightly outperforming the baseline control model at \(17.44\%\). In contrast, the Syria corridor achieved its strongest performance under Direct LLM-Lasso, which reduced forecasting error from \(31.44\%\) in the control configuration to \(29.29\%\). The Ukraine corridor achieved its lowest forecasting error under the proposed LLM-Assisted Regularization (AR) framework, which produced a MAPE of \(41.05\%\) compared to \(61.66\%\) for the baseline gravity model.

\begin{figure}[t]
    \centering
    \includegraphics[width=0.98\columnwidth]{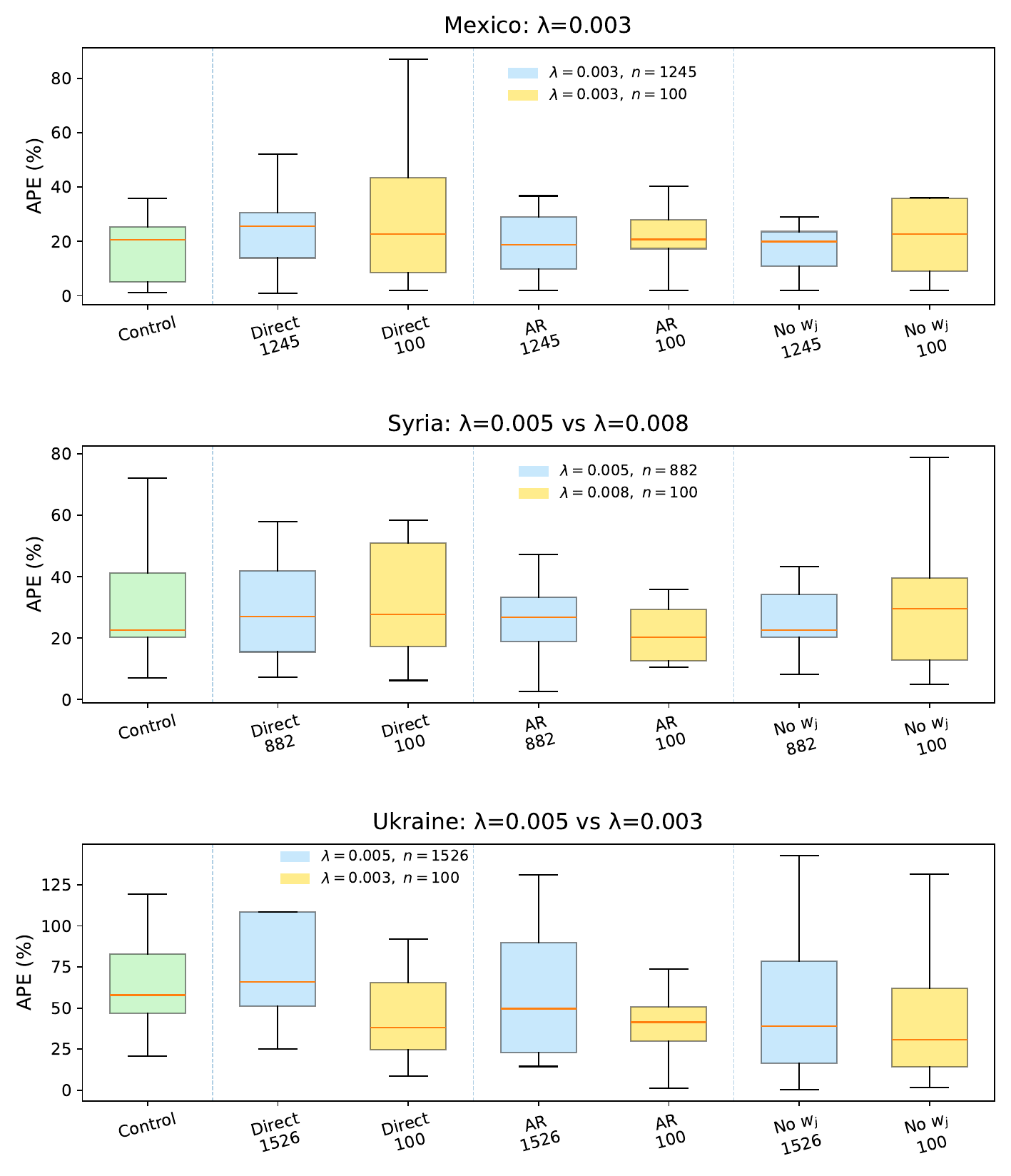}
    \caption{Box plot comparison of absolute percentage error distributions across migration forecasting models for Mexico, Syria, and Ukraine under varying article counts and regularization settings. Lower values indicate improved predictive accuracy.}
    \label{fig:combined_boxplots}
\end{figure}

Figure~\ref{fig:combined_boxplots} illustrates the distribution of absolute percentage error (APE) values across forecasting periods for the three migration corridors. The box plots provide additional insight into forecasting stability and variance beyond aggregate MAPE values alone.

Across several evaluated configurations, Direct LLM-Lasso and AR demonstrated greater sensitivity to article-corpus size than the baseline gravity approaches. Larger article corpora did not consistently reduce forecasting error and frequently produced greater variance across forecasting periods, particularly for the Ukraine corridor. In contrast, several smaller article configurations produced narrower forecasting-error distributions and lower median APE values. Direct LLM-Lasso frequently achieved forecasting performance comparable to AR, but generally exhibited wider distributions and greater variance across forecasting periods. Because each migration corridor contains a relatively small number of monthly holdout forecasts, modest forecasting differences between configurations may partially reflect temporal forecasting noise and article-content variability rather than definitive superiority of a particular modeling strategy. Consequently, the observed performance differences should be interpreted primarily as exploratory comparative results rather than statistically conclusive outcomes.

\section{Conclusion}

This work investigated whether LLM-assisted contextual information extracted from news media could improve migration forecasting within gravity-based forecasting frameworks. Rather than using LLMs as direct migration predictors, the proposed framework incorporated article-derived signals into weighted Lasso regularization and push--pull feature construction.

Experimental results demonstrated that hierarchical inference pipelines combining lightweight URL-level analysis with selective contextual escalation were capable of extracting migration-related push--pull signals from unstructured news data. Across the evaluated temperature settings, classification behavior remained relatively stable, suggesting that the constrained prompt structure reduced temperature sensitivity. Among the evaluated configurations, the proposed SuLuSa hierarchy consistently produced the strongest overall classification behavior by combining lightweight inference, larger-model reasoning, and full-text contextual analysis within a multi-stage framework.

The forecasting experiments demonstrated mixed performance across migration corridors and forecasting configurations, with no single regularization strategy consistently outperforming all competing approaches. While Direct LLM-Lasso occasionally improved forecasting performance relative to the baseline gravity models, the proposed LLM-Assisted Regularization (AR) framework generally produced more stable behavior across several evaluated configurations. Results additionally showed that increasing article volume alone did not consistently improve forecasting accuracy and frequently introduced additional variance and instability, particularly under larger article-corpus configurations.

Analysis of forecasting error distributions further suggested that AR often produced lower-variance and more stable forecasting behavior under smaller article configurations, while Direct LLM-Lasso generally exhibited noisier behavior and greater sensitivity to outlier forecasting periods. These findings suggest that contextual signal quality, article relevance, and migration-corridor characteristics may play a more important role than raw article quantity alone.

Overall, this work demonstrates that unstructured news data can improve migration forecasting when incorporated indirectly through contextual signal extraction and structured feature weighting. At the same time, the results highlight important limitations associated with directly generated LLM-based regularization penalties and emphasize the importance of stability, interpretability, contextual consistency, and article quality in LLM-assisted forecasting frameworks.

\bibliographystyle{IEEEtran}
\bibliography{example}

\appendices
\section{URL-Level Classification Prompt}
\label{appendix:URL-Level Classification Prompt}
The following prompt was used for lightweight URL-level migration-pressure classification within the hierarchical inference pipeline.

\begin{lstlisting}
You are a URL-only migration-pressure classifier.
Use ONLY lexical tokens found in the URL string. No outside knowledge. No invented details.

Country-level interpretation:
Assume the article concerns migration pressure affecting the country referenced in the URL.
- push: signals conditions that may cause people to LEAVE that country
- pull: signals conditions that may cause people to ARRIVE in that country

Interpret pressure relative to the country mentioned or implied by the URL.

Goal:
For each URL, classify migration pressure:
- push: conditions encouraging emigration (war, crisis, crackdown,
  collapse, unemployment, disaster, displacement)
- pull: conditions encouraging immigration (jobs, hiring, growth,
  investment, visa, asylum, residency, aid, safety)
- mixed: both push and pull indicators appear
- neutral: no clear migration-related tokens
- unclear: URL too short, generic, ID-like, or ambiguous

Language rule:
Reasons MUST be written in English even if the URL tokens are non-English.
Quote the exact token(s) from the URL that triggered the classification.

Output (STRICT JSONL, one object per line, no extra text):
{"row_id": <int>, "pressure": "push|pull|mixed|neutral|unclear",
 "reason": "<=12 English words; include exact URL token(s)"}

ITEMS (tab-separated):
<row_id>\t<url>
\end{lstlisting}

\section{LLM Prompt for Article-Based Push--Pull Classification}
\label{appendix:LLM Prompt for Article-Based Push--Pull Classification}
The following prompt was used during the article-level
migration-pressure classification stage of the SuLuSa pipeline.

\begin{lstlisting}
You are an article-based migration-pressure classifier.

Use ONLY the provided article text. No outside knowledge.
No invented details.

Country-level interpretation:
Classify migration pressure relative to the main country
discussed in the article.

- push: conditions that may cause people to LEAVE that country
- pull: conditions that may cause people to ARRIVE in that country

Goal:
Classify the article as one of:
- push
- pull
- mixed
- neutral
- unclear

Definitions:
- push: war, violence, persecution, repression,
  unemployment, collapse, disaster, displacement
- pull: jobs, hiring, growth, investment, visas,
  asylum access, residency, aid, safety, stability
- mixed: both push and pull are meaningfully present
- neutral: article is not clearly about
  migration-driving conditions
- unclear: article text is too vague or insufficient

Language rule:
Reason MUST be written in English.

Output (STRICT JSON ONLY, no markdown, no extra text):
{"pressure":"push|pull|mixed|neutral|unclear",
 "reason":"<=16 English words"}

ARTICLE:
<article text inserted here>
\end{lstlisting}

%

\section{Feature-Category Counting Prompt for LLM-Assisted Regularization}
\label{appendix:Feature-Category Counting Prompt for LLM-Assisted Regularization}
The following prompt was used for counting features 

\begin{lstlisting}
You are counting migration-model feature evidence in this {source_type}.

Task:
For each feature, count how many distinct pieces of evidence in the text relate to that feature.

Important:
- Count SEMANTIC evidence, not keyword frequency.
- Use contextual meaning and overall article content when determining relevance.
- The listed keywords are conceptual examples intended to describe each feature category
- Do NOT mechanically search for keywords.
- Count semantic evidence.
- Count distinct claims, events, or conditions.
- Evidence may support a feature even if none of the exact keywords appear.
- Features are NOT mutually exclusive.
- If one sentence supports multiple features, count each relevant feature.
- Overlapping evidence should contribute to multiple categories when appropriate.
- Do NOT force evidence into only one category.
- Count distinct claims, events, situations, or conditions.
- If a feature is not supported, return 0.
- Use ONLY the provided text.
- Return ONLY valid JSON.

Feature meanings and contextual examples:

- gpi(global peace index):
  conflict, war, violence, insecurity, instability, persecution,
  displacement, refugees, safety, peace, military activity

- hdi(human development index):
  human development, living standards, poverty, welfare,
  quality of life, inequality, wellbeing, development conditions

- life_expectancy:
  health, mortality, hospitals, disease, epidemics,
  public health, deaths, healthcare strain

- expected_schooling:
  schools, education, students, teachers,
  universities, classroom access, learning disruption

- gdp_per_capita:
  economy, jobs, income, wages, inflation,
  unemployment, recession, growth, trade, business activity

- distance:
  borders, crossings, migration routes, travel,
  neighboring countries, migration corridors, journey difficulty

Text:
{text[:3000]}

Return ONLY valid JSON using this exact structure.
Replace the example numbers with your actual semantic evidence counts.

{
  "feature_counts": {
    "gpi": 0,
    "hdi": 0,
    "life_expectancy": 0,
    "expected_schooling": 0,
    "gdp_per_capita": 0,
    "distance": 0
  }
}
""".strip()
\end{lstlisting}

\section{Direct LLM-Lasso Prompt}
\label{appendix:Direct LLM-Lasso Prompt}
The following prompt was used in the Direct LLM-Lasso framework to assign feature-specific monthly regularization penalties directly from article evidence. The language model was instructed to estimate relative feature importance for each month using contextual information extracted from the corresponding article corpus.

\scriptsize
\begin{lstlisting}
You are helping build a migration-flow prediction model using LLM-Lasso.

Your task is to directly assign monthly regularization penalties w_j
for each base feature using ONLY the article evidence from month {month}.

Meaning of w_j:
- w_j must be between 0.1 and 1.0.
- Lower w_j = feature is more important this month and should
  be penalized LESS.
- Higher w_j = feature is less important this month and should
  be penalized MORE.

Suggested scale:
- 0.1 to 0.3 = highly important / strongly supported
  by the month evidence
- 0.4 to 0.7 = moderately important / partially supported
- 0.8 to 1.0 = weakly important / little support

Rules:
- Return one score for EVERY feature listed below.
- Use the articles as evidence, but you may use
  migration-domain reasoning.
- Features do not need to be literally named to be important.
- Do NOT use mention counts or 1/count logic.
- Do NOT assign all features the same penalty.
- Be comparative: decide which features mattered more
  in this month.
- Use valid JSON only. No markdown fences.

Features:
{feature_block}

Article evidence for {month}:
{article_context}

Required output structure:
{
  "feature_name": 0.5
}
\end{lstlisting}
\normalsize




\section{Temperature sensitivity results}\label{app:temperature}

Table~\ref{tab:temperature} presents the results of the temperature sensitivity experiments, showing the number of articles classified into each migration-related category across the evaluated inference pipelines. Overall, temperature variation had relatively little effect on classification behavior, with the distributions of push, pull, mixed, and unclear classifications remaining broadly stable across all evaluated settings. For example, the Qwen SuLuSa pipeline classified 163 articles as ``unclear'' at a temperature of $\tau = 0.3$, compared to 159 articles at $\tau = 1.5$, demonstrating only minor variation despite the substantially higher level of generation randomness. Similar patterns were observed across the other evaluated inference hierarchies, suggesting that the structured prompt design constrained the range of possible outputs even when higher temperature values were used.

Although higher temperatures occasionally produced slightly larger numbers of ``mixed'' classifications, the overall classification distributions remained highly consistent across runs. This relative stability likely occurred because the models were required to follow a constrained classification schema consisting of predefined push, pull, mixed, neutral, and unclear categories rather than generating unrestricted free-form responses. As a result, even at higher temperature settings, the inference pipelines still operated within the boundaries imposed by the prompt structure and output requirements. These findings suggest that the hierarchical inference configuration itself contributed more to classification consistency and overall performance than temperature selection alone.

\begin{table}[tbp]
\centering
\caption{Temperature sensitivity results.  }
\label{tab:temperature}

\setlength{\tabcolsep}{3pt}
\begin{tabular}{cc|ccccc}
\toprule
Temp. & Mode & Push & Pull & Mixed & Neutral & Unclear \\
\midrule
\multirow{4}{*}{0.3} & Qwen SuSa & 152 & 27 & 2 & 0 & 200 \\
 & Qwen SuLu & 124 & 46 & 0 & 0 & 211 \\
 & Qwen SuLuSa & 164 & 54 & 0 & 0 & \textbf{163}\\
 & Mistral SuSa & 160 & 42 & 4 & 0 & 168 \\
\midrule
\multirow{4}{*}{0.8}
 & Qwen SuSa & 162 & 24 & 0 & 0 & 195 \\
 & Qwen SuLu & 126 & 35 & 1 & 0 & 219 \\
 & Qwen SuLuSa & 157 & 60 & 1 & 0 & \textbf{163} \\
 & Mistral SuSa & 150 & 40 & 15 & 0 & 172 \\
\midrule
\multirow{4}{*}{1.5}
 & Qwen SuSa & 148 & 27 & 1 & 0 & 205 \\
 & Qwen SuLu & 115 & 40 & 3 & 0 & 223 \\
 & Qwen SuLuSa & 175 & 41 & 6 & 0 & \textbf{159} \\
 & Mistral SuSa & 142 & 45 & 14 & 0 & 173 \\
\bottomrule
\end{tabular}
\end{table}

\section{SuLuSa Across Temperatures}

Table~\ref{tab:temp_sensitivity} demonstrates that the SuLuSa pipeline remained relatively consistent both across repeated runs and across evaluated temperature settings ranging from $\tau = 0.3$ to $\tau = 1.5$. SuLuSa consistently achieved the strongest overall performance among the evaluated inference hierarchies, likely because it incorporated all three progressive inference stages (Su, Lu, and Sa), enabling increasingly deeper contextual analysis when classifying push, pull, mixed, and unclear migration-related signals. Although minor classification variability was observed at higher temperature settings, the overall distribution of classifications remained broadly stable across runs.

\begin{table}[tbp]
\centering
\setlength{\tabcolsep}{3pt}
\small
\caption{Inference Performance for mode \textbf{SuLuSa} across temperatures}
\label{tab:temp_sensitivity}
\begin{tabular}{cccccc}
\toprule
Temperature & Push & Pull & Mixed & Neutral & Unclear \\
\midrule
0.1 & 165 & 55 & 1 & 0 & 160 \\
0.3 & 164 & 54 & 0 & 0 & 163 \\
0.5 & 173 & 46 & 4 & 0 & 158 \\
0.8 & 157 & 60 & 1 & 0 & 163 \\
0.9 & 159 & 48 & 2 & 0 & 170 \\
1.5 & 175 & 41 & 6 & 0 & 159 \\
\bottomrule
\end{tabular}

\end{table}

%

\section{Article Retrieval Queries}
\label{appendix:article_queries}

Migration-related articles were collected using source-filtered
DuckDuckGo Search (DDGS) queries combining migration-corridor
region terms, migration-related event keywords, month identifiers,
and source-specific domain filters. Queries were executed separately
for each migration corridor, news source, and monthly forecasting
window spanning November 2021 through November 2022.

The following generalized query templates were used throughout
the article retrieval process:

\begin{lstlisting}
site:{domain} {region_pair} {event_pair} {month} {year}

site:{domain} {region_pair} {event_main} {year}

site:{domain} {region_main} {event_main} {month} {year}

site:{domain} {region_main} news {event_main} {year}
\end{lstlisting}

The following migration-corridor terms were used:

\begin{itemize}
    \item \textbf{Ukraine--Poland:}
    region terms = \{Ukraine, Russia\},
    event terms = \{war, conflict, invasion, refugees\}

    \item \textbf{Mexico--United States:}
    region terms = \{Mexico, United States, US border\},
    event terms = \{migration, immigration, asylum, migrants\}

    \item \textbf{Syria--Turkey:}
    region terms = \{Syria, Turkey, Syrian\},
    event terms = \{civil war, conflict, refugees, displacement\}
\end{itemize}

The following news domains were queried:

\begin{itemize}
    \item bbc.com
    \item npr.org
    \item pbs.org
    \item nbcnews.com
    \item cbsnews.com
    \item reuters.com
    \item apnews.com
    \item theguardian.com
    \item nytimes.com
    \item washingtonpost.com
    \item abcnews.go.com
\end{itemize}

An example query generated for the Ukraine migration corridor
is shown below:

\begin{lstlisting}
site:bbc.com Ukraine Russia war conflict November 2021
\end{lstlisting}

\section{Best Models For Country Pair}
\label{appendix:Best_Model_for_each_country_Pair}

Table~\ref{tab:exp5_results} summarizes the best-performing forecasting configurations identified for each evaluated migration corridor. Configurations were compared across forecasting method, $\lambda$ regularization value, and article corpus size. 

The evaluated article corpus sizes correspond to independently pre-constructed migration-news datasets generated during the scraping stage. For example, configurations such as ``Ukraine\_100'' or ``Mexico\_1245'' refer to forecasting experiments performed using CSV datasets containing different quantities of migration-related news articles collected across the evaluated temporal window. These article corpora were loaded directly into the forecasting pipeline during experimentation rather than dynamically generated during model training.

Results indicate that the strongest-performing forecasting configuration varied across migration corridors. The Mexico--US corridor achieved its lowest forecasting error using the No $w_j$ configuration, while the Ukraine--Poland corridor performed best using the proposed LLM-Assisted Regularization (AR) framework. In contrast, the Syria--Turkey corridor achieved its strongest performance using the Direct LLM-Lasso approach.



\begin{table}[tbp]
\centering

\caption{Best-performing forecasting configurations across migration corridors.}
\label{tab:exp5_results}

\begin{tabular}{l l c c}
\toprule
\multicolumn{4}{c}{\textbf{Forecasting Results}} \\
\midrule
\textbf{Migration Corridor} & \textbf{Best Model} & $\boldsymbol{\lambda}$ & \textbf{$N$ Articles} \\
\midrule
Mexico & No $w_j$ & 0.003 & 1,245 \\
Ukraine & LLM-Assisted Reg. (AR) & 0.003 & 100 \\
Syria & Direct Lasso & 0.005 & 882 \\
\bottomrule
\end{tabular}

\end{table}

\section{Lambda Values}
\label{app:top_configs}
Table~\ref{tab:lambda_values} summarizes the Lasso regularization strengths evaluated throughout the forecasting experiments using the proposed SuLuSa hierarchical inference framework. Multiple $\lambda$ values were tested in order to examine how different levels of regularization influenced migration-flow forecasting performance across the evaluated migration corridors. These experiments allowed for comparison between weaker and stronger regularization settings under identical forecasting conditions. The selected $\lambda$ values were subsequently incorporated into the rolling forecasting experiments discussed in later sections.

\begin{table}[tbp]
\centering
\caption{Evaluated Lasso regularization strengths used throughout forecasting experiments.}
\label{tab:lambda_values}

\begin{tabular}{l l}
\toprule
\textbf{Hyper-Parameter} & \textbf{Evaluated Values} \\
\midrule

Lasso Regularization Strength ($\lambda$)
&
$7\times10^{-4}$,
$9\times10^{-4}$,
$1\times10^{-3}$,
\\
&
$3\times10^{-3}$,
$5\times10^{-3}$,
$6\times10^{-3}$,
\\
& 
$7\times10^{-3}$,
$8\times10^{-3}$
\\
Primary Experimental Values
&
$\{1,3,5,8\}\times10^{-3}$
\\

\bottomrule
\end{tabular}
\end{table}
\end{document}